\documentclass{article}
\usepackage{spconf,amsmath,graphicx}
\usepackage{comment}
\usepackage{amssymb}
\usepackage{subfig,fixltx2e,url}

\graphicspath{ {./images/} }

\title{Diversity in Fashion Recommendation using Semantic Parsing}
\name{Sagar Verma$^{1}$ \qquad Sukhad Anand$^{1}$ \qquad Chetan Arora$^{1}$ \qquad Atul Rai$^{2}$}

\address{$^{1}$IIIT Delhi \qquad $^{2}$Staqu Technologies}

\begin{document}
%
\maketitle
\begin{abstract}
Developing recommendation system for fashion images is challenging due to the inherent ambiguity associated with what criterion a user is looking at. Suggesting multiple images where each output image is similar to the query image on the basis of a different feature or part is one way to mitigate the problem. Existing works for fashion recommendation have used Siamese or Triplet network to learn features between a similar pair and a similar-dissimilar triplet respectively. However, these methods do not provide basic information such as, how two clothing images are similar, or which parts present in the two images make them similar. In this paper, we propose to recommend images by explicitly learning and exploiting part based similarity. We propose a novel approach of learning discriminative features from weakly-supervised data by using visual attention over the parts and a texture encoding network. We show that the learned features surpass the state-of-the-art in retrieval task on DeepFashion dataset. We then use the proposed model to recommend fashion images having an explicit variation with respect to similarity of any of the parts.
\end{abstract}

\begin{keywords}
Fashion Recommendation, Semantic Parsing of Clothing Items
\end{keywords}

\section{Introduction}
\label{sec:intro}

The recent advancements in computer vision and machine learning techniques have made it possible to match images and objects captured under different pose and environmental conditions. One popular application of such a system is in online shopping for clothing items, where researchers have used the technique for product retrieval, fashion recommendation, style recognition, and clothing item parsing, etc. The image-based recommendation is a crucial part of any fashion e-commerce website which primarily undertakes two kinds of retrieval tasks; in-shop and consumer-to-shop retrieval. In-shop retrieval task requires searching for similar product images, all captured in a shop environment, whereas consumer-to-shop retrieval task requires matching shop images that are similar to user-shot query images. Understandably, the later problem is much harder due to uncontrolled lighting conditions, body pose and object occlusion in the user-shot images.

\begin{figure}[t]
\includegraphics[width=1.0\linewidth]{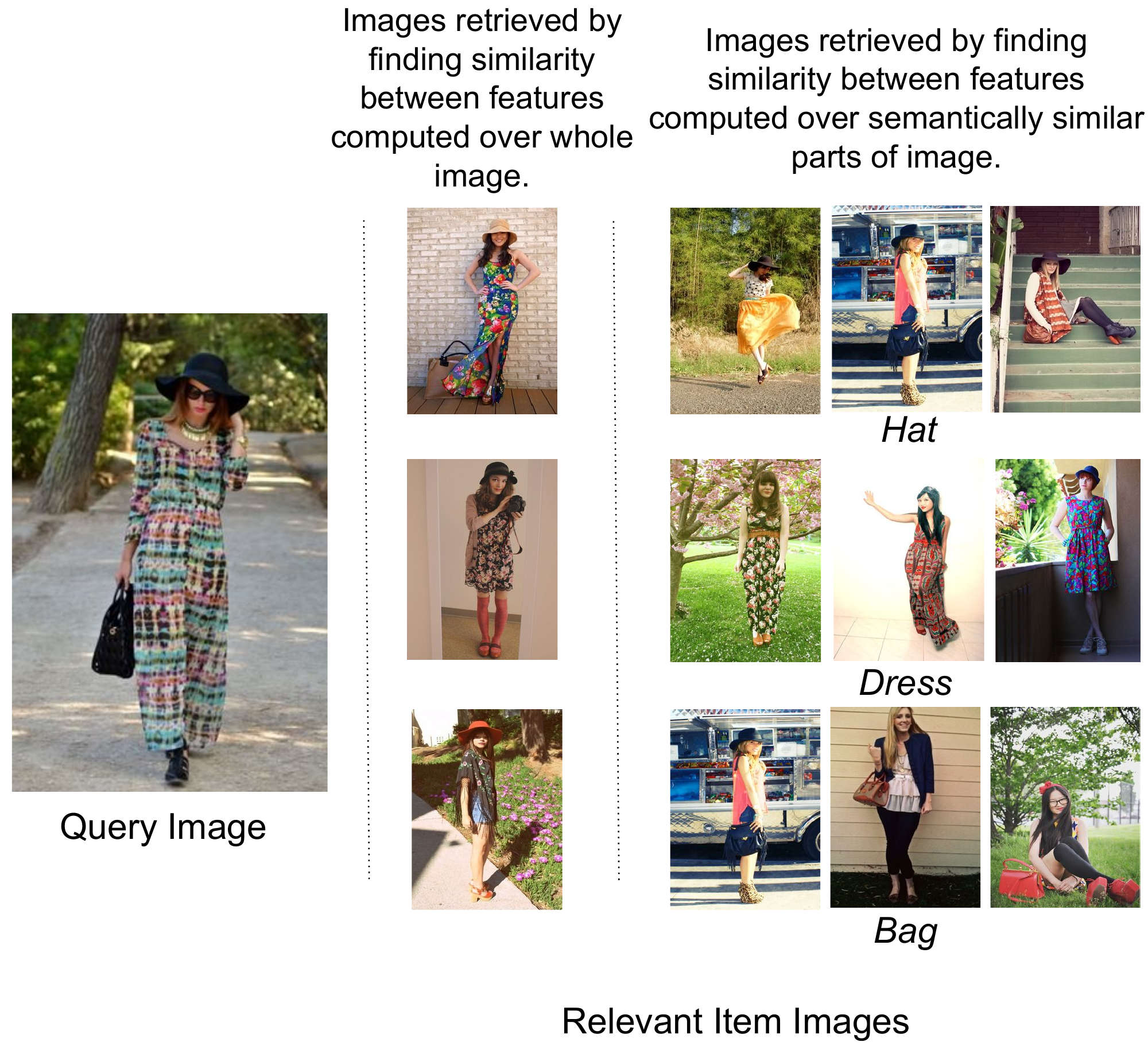}
\caption{Finding similarity based on the visual features computed over the whole image gives results that look similar because of some large clothing item present in the images as shown on the middle column. Proposed work recognizes the distinct and fine clothing items present in the query image and then finds similar images to those items category wise as show on right column.}
\label{fig:similar_items}
\end{figure}

Existing methods for fashion recommendation focus on finding similarity between two images using Siamese network \cite{HuangICCV2015, VeitICCV2015, WangICMR2016}. In an extension of the work, it is also possible to learn similarity-dissimilarity between three images using a Triplet network \cite{WangCVPR2014, SimoSerraCVPR2016}. Both Siamese, and Triplet Networks typically focus on the whole image to learn visual features and usually capture only the dominant visual cue. This becomes problematic, especially in the case of consumer-to-shop fashion recommendation tasks, since the pose of the person in query image is uncontrolled. This makes it difficult to infer, which part or clothing item the user is interested in. The problem becomes even more challenging, if the desired item is visually smaller, making it harder for convolutional neural networks, a technique of choice in many such systems, to focus on such non-dominant cues.

To handle inherent ambiguity in user's intentions, in this paper, we propose a novel fashion recommendation system, which implicitly hedges its suggestions on the basis of similarity with various parts in the query image. We propose to use attention module in deep neural networks to focus on distinct parts of an image and retrieve similar products for each part based on the texture similarity. We treat the problem as multi-label classification and use a single convolutional stream to learn visual features and an attentional LSTM to focus on different locations. Our convolutional stream contains a texture encoding layer in the end to learn texture based similarity from each attention region.

The specific contributions of this work are as follows :
\begin{itemize}
\item Given the ambiguous nature of user intent, we propose a new recommendation system to produce diverse recommendations on the basis of similarity of different parts in the query image.
\item To generate semantically meaningful parts in the fashion image, we propose to use attention based deep neural network which learns to attend to different parts using the weakly labeled data available in the benchmark dataset.
\item Instead of features from standard pre-trained neural networks, we suggest using texture-based features which, as we show in our experiments, are better suited for finding clothing similarity.
\item Apart from diversity in fashion recommendation, our experiments and evaluations on multiple datasets demonstrate the superiority of the proposed model in attribute classification and cross-scene image retrieval tasks.
\end{itemize}

\section{Related Work}

In the past few years there has been a tremendous increase in the research on fashion and related domains \cite{SimoSerraACCV2014, YamaguchiCVPR2012} such as cloth parsing \cite{YamaguchiPAMI2015, yang2014clothing}, and clothing attribute recognition \cite{chen2012describing, YamaguchiBMVC2015}. Detecting fashion style \cite{VeitICCV2015} and more subjective attribute such as ``how fashionable a person looks" have also been looked at \cite{SimoSerraCVPR2015}. A task closely related to our goal is cross-domain image retrieval \cite{LiuCVPR2012, HuangICCV2015}, where most of the current techniques have focussed on the whole image for computing similarity score. Wang et. al. \cite{WangVCIP2017} used attention for the task, whereas Chen et. al. \cite{ChenAAAI2018} have shown that the visual attention architecture of image captioning task cannot be used for multi-label classification as no sequence information is present. Wang et. al. \cite{WangSTLSTMICCV2017} have proposed a similar architecture, as used in this paper, using spatial transformer network and hard attention.

In this paper, we propose to learn semantically important regions and then generate diverse recommendations on the basis of similarity of any of these. We note that most of the state-of-the-art learning systems for image matching, are geared towards global shape matching. The networks learn to be invariant against the fine variations and distortions. However, in case of fashion and clothing, the fine-grained differences in style and texture are the most important cues. Therefore we propose to match the recovered semantic regions on the basis of texture rather than standard features obtained from ResNet \cite{resnet50} or VGGNet \cite{vgg2014}. Texture has been a well-studied problem in computer vision where both traditional hand tuned \cite{LokeICCE2017, LiuMM2012} as well as deep learned features \cite{cimpoi2015deep, andrearczyk2016using} have been proposed. We use the texture encoding layer proposed by Zhang et. al. \cite{ZhangCVPR2017} in our architecture. The encoding layer generalizes robust residual encoders such as VLAD \cite{VLADCVPR2010} and Fisher Vectors \cite{FisherCVPR2007} and can be trained end to end, along with the rest of the model.

\section{Architecture}

\begin{figure}[t]
\centering
\includegraphics[width=1.0\linewidth]{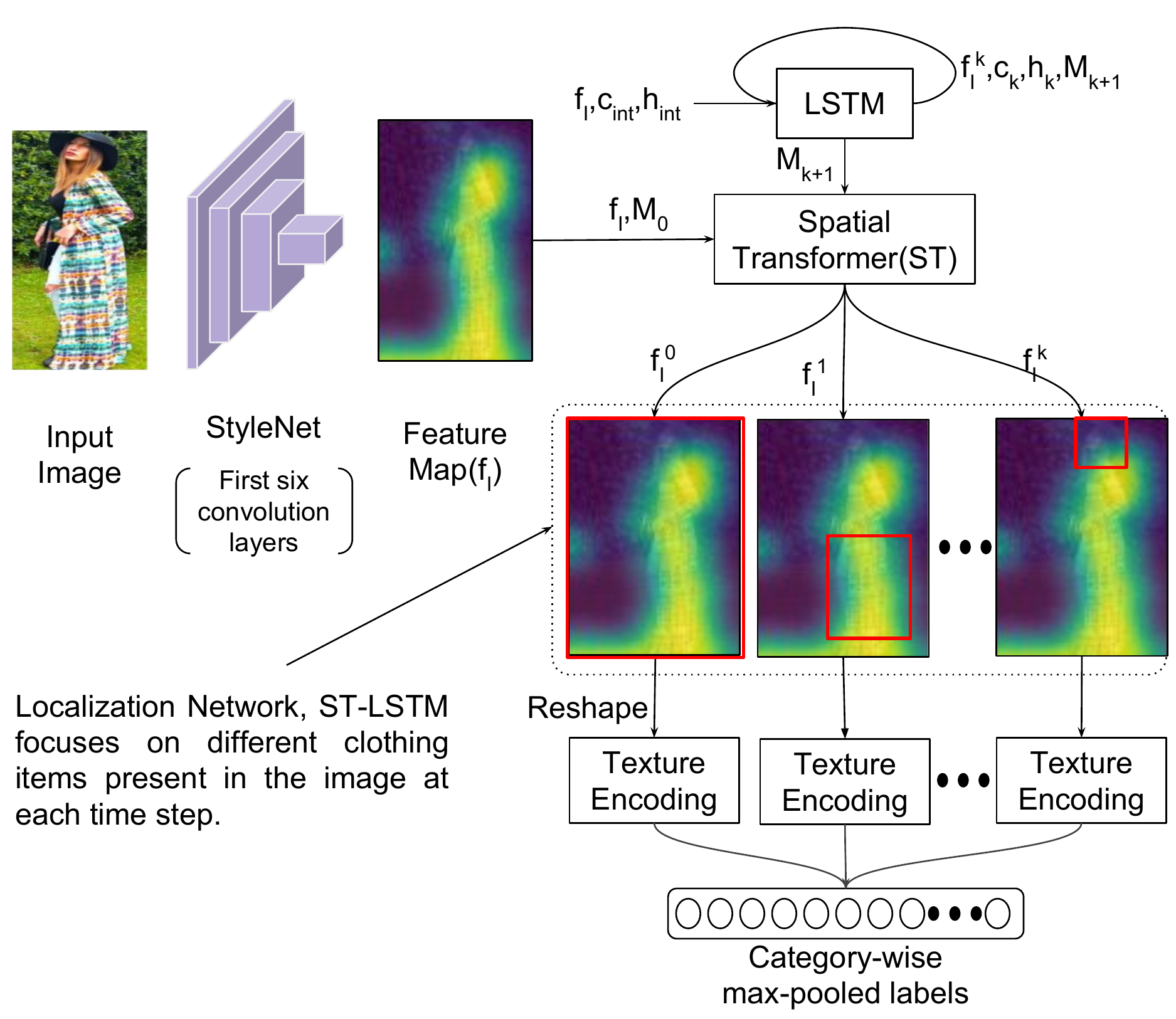}
\caption{Proposed architecture}
\label{fig:res}
\end{figure}

Our proposed architecture consists, a six-layered convolutional neural network to extract global image features, a visual attention module to generate attention over an image, and a texture encoding layer to capture clothing texture from the generated attentions. Since we would like the proposed model to capture similarity of various semantic regions in the image, we first train the model to explicitly learn these important regions. We do so by using an attention module, which learns to focus on and predict different objects present in the image. Owing to the space restriction, we briefly describe below, important aspects of the proposed architecture. More details can be found at the project page: \url{www.iiitd.edu.in/~chetan/projects/fashion/}. 

\paragraph*{CNN for Global Image Features:}

We use the first three convolutional and max-pooling layers of the model used in \cite{SimoSerraCVPR2016} to extract global image features. The model has been trained specifically for clothing and fashion images and worked best in our experiments. We note that other similar architectures could have been used as well.

\paragraph*{Visual Attention Module: }

For attention we use a recurrent memorized-attention module, which combines the recurrent computation process of an LSTM network, a spatial transformer\cite{JaderbergNIPS2015} and a texture encoding layer \cite{ZhangCVPR2017} described in the next section. It iteratively searches different parts and predicts the scores of label distribution for them and also the corresponding feature vectors.

\paragraph*{Texture Encoding Layer: }

The features extracted at each time step of the LSTM network are passed through a texture encoding layer which learns the texture of the attentional region. The encoding layer incorporates dictionary learning, feature pooling and classifier learning into it. The layer assumes $k$ codewords and assigns descriptors to them using soft assignment so that the function is differentiable and can be trained along with the model using Stochastic gradient descent with backpropagation as proposed by Zhang et. al.\cite{ZhangCVPR2017}.

\paragraph*{Training: }

Given a dataset with $N$ training sample images and $C$ class labels, during training each image is passed through the complete network. Multiple labels for an input image is represented as one-hot-vector with 1 corresponding to all the labels present. A CNN generates global spatial features which are fed into the attention module which further generates localized texture features of the attentional region and corresponding labels. The LSTM network is unrolled for $k$ time steps and hence produce $k$ labels by category wise max pooling.

\paragraph*{Loss: }

We use Euclidean distance as the objective function for classification loss as given in \cite{WangSTLSTMICCV2017}. In order to force localization network to look at a different location at each time step, we use divergence loss given in \cite{ZhaoToP2017}. Divergence loss is computed by finding the correlation between two subsequent attention maps. We also use localization loss given in \cite{WangSTLSTMICCV2017} which removes redundant locations and forces localization network to look at small clothing parts. All three loss values are added together with multiplicative factors of $1$ for classification loss and localization loss and $0.01$ for divergence loss for all of our experiments.

\begin{table}
\centering
\begin{tabular}{ l c c c c}
 \hline
Dataset & \multicolumn{2}{c}{Fashion144k \cite{SimoSerraCVPR2016}} & \multicolumn{2}{c}{Fashion550k \cite{InoueICCVW2017}} \\
Model & \textit{AP}\textsubscript{all} & \textit{mAP} & \textit{AP}\textsubscript{all} & \textit{mAP} \\
\hline
StyleNet \cite{SimoSerraCVPR2016} & 65.6 & 48.34 & 69.53 & 53.24 \\
Baseline \cite{InoueICCVW2017} & 62.23 & 49.66 & 69.18 & 58.68 \\
Viet et al. \cite{veit2017learning} & NA & NA & 78.92 & 63.08 \\
Inoue et al. \cite{InoueICCVW2017} & NA & NA & 79.87 & \textbf{64.62} \\
Ours  & \textbf{82.78} & \textbf{68.38} & \textbf{82.81} & 57.93 \\
 \hline
\end{tabular}
\caption{Multi-label classification on Fashion144k \cite{SimoSerraCVPR2016} and Fashion550k \cite{InoueICCVW2017}}
\label{table:multi-label}
\end{table}

\begin{figure}%
    \centering
    \subfloat[In-Shop retrieval]{{\includegraphics[width=0.48\linewidth]{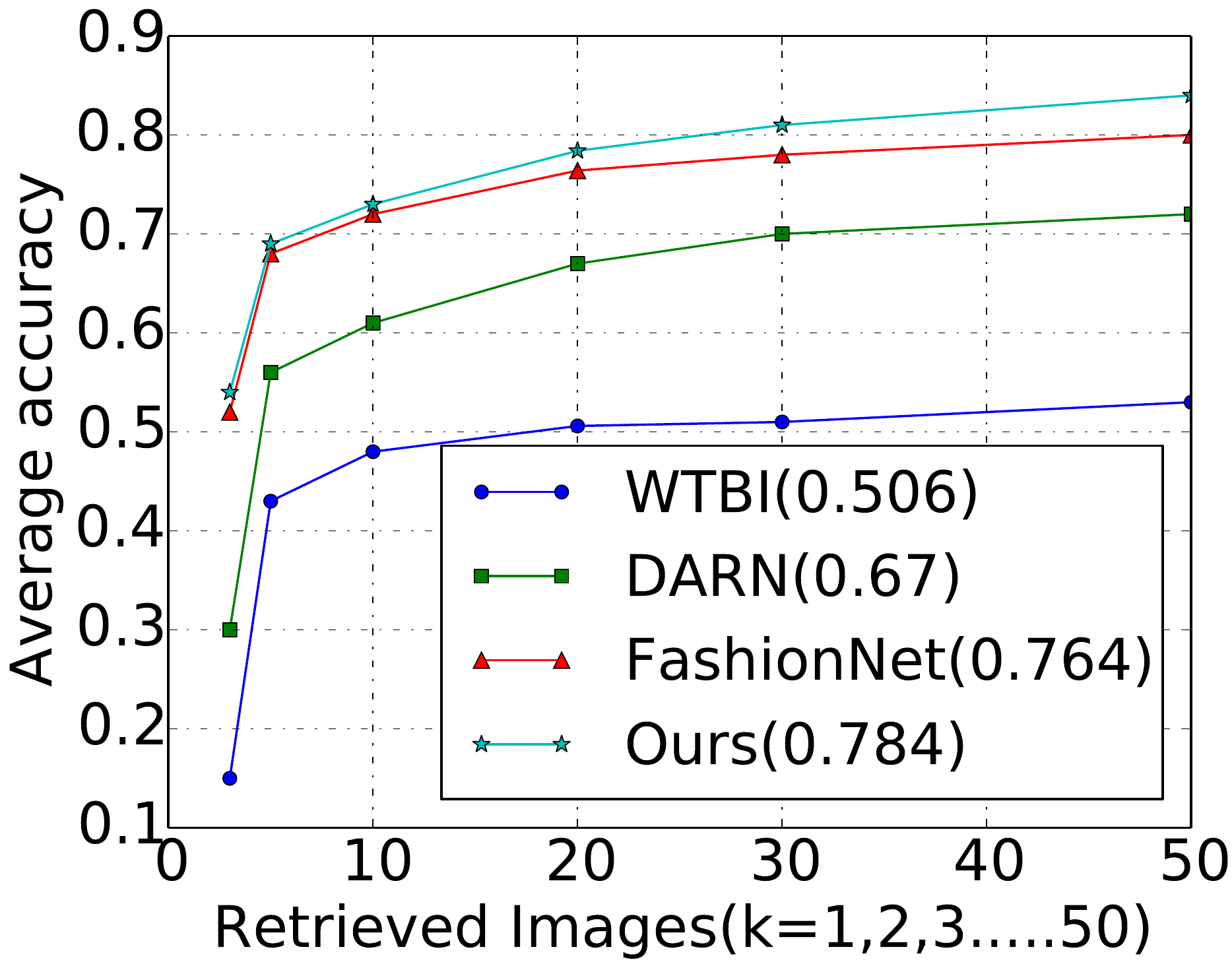} }}
    \subfloat[Consumer-to-shop retrieval]{{\includegraphics[width=0.48\linewidth]{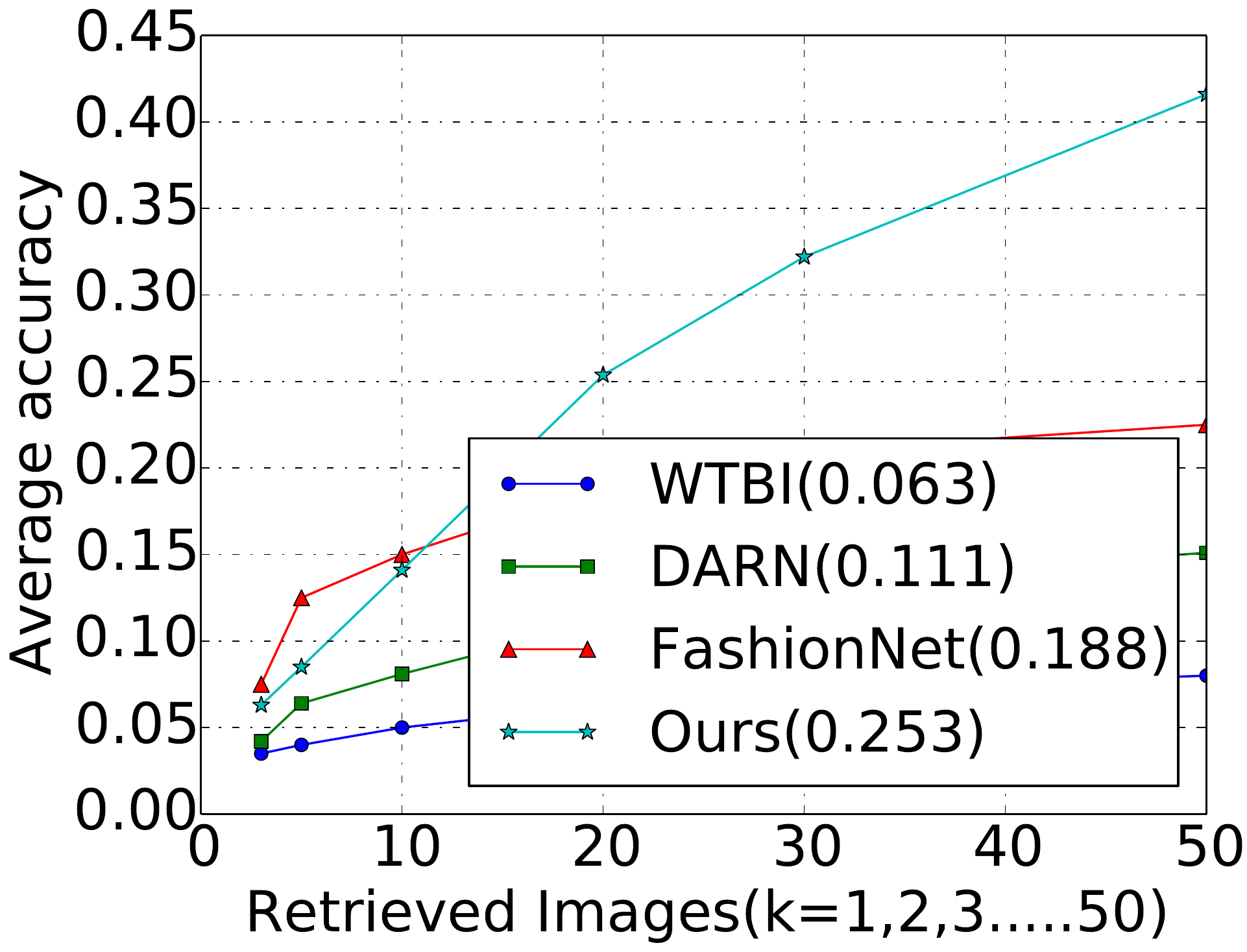} }}
    \caption{Retrieval results for In-shop and Consumer-to-shop retrieval tasks on DeepFashion dataset \cite{LiuCVPR2016}. Proposed method works better in both the tasks.}
    \label{fig:retrieval_results}
\end{figure}

\begin{figure*}[t]
\centering
\includegraphics[width=1.0\linewidth]{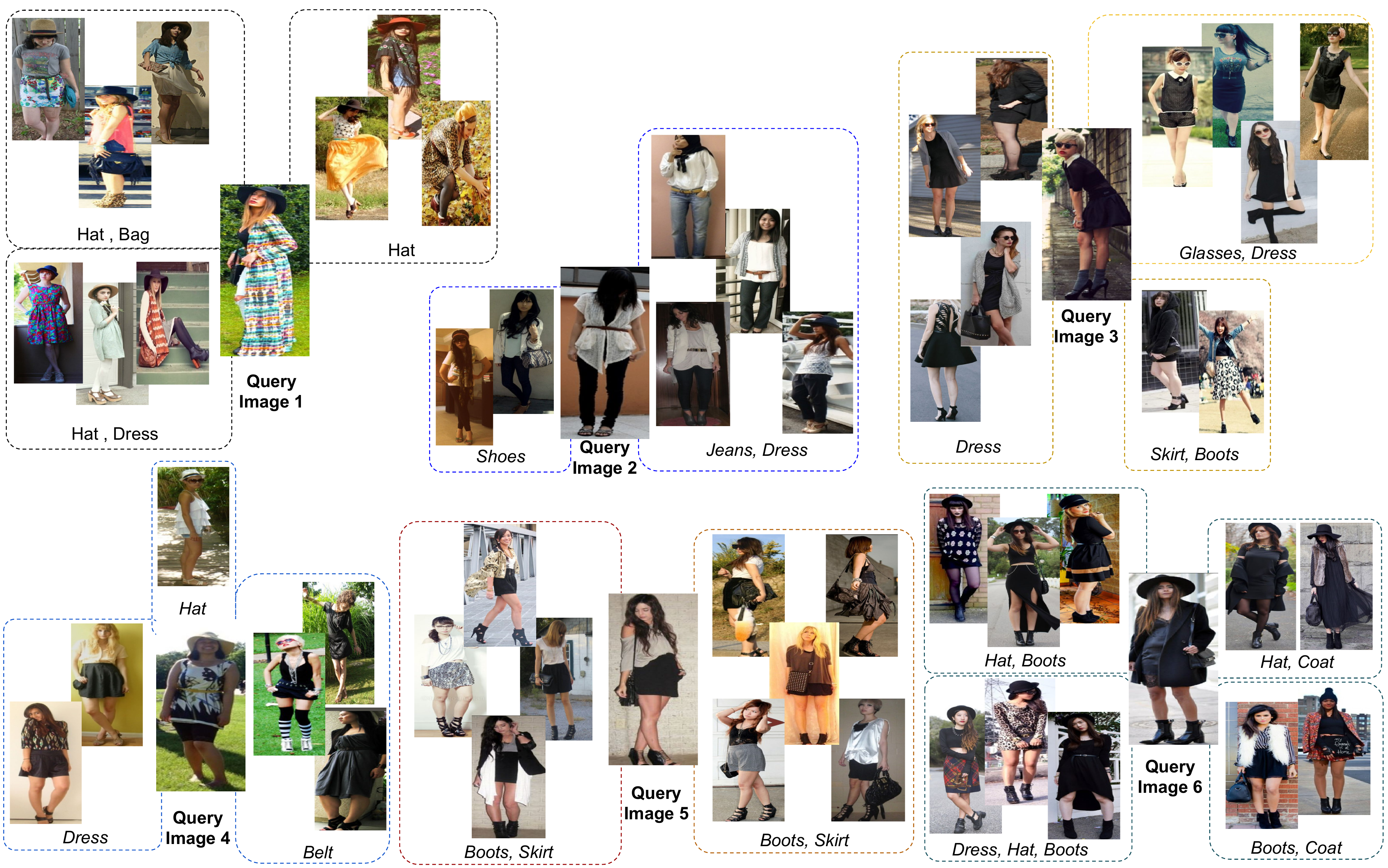}
\caption{Semantically similar results for some of the query images from Fashion144k dataset \cite{SimoSerraCVPR2016} using our method. Retrieved images for six query images are shown from left to right, top to bottom. For each query image, retrieved images having similar clothing items are shown together.}
\label{fig:results}
\end{figure*}

\section{Datasets}

We have conducted our experiments on Fashion144K \cite{SimoSerraCVPR2016}, Fashion550k \cite{InoueICCVW2017}, and DeepFashion dataset \cite{LiuCVPR2016}.

The cleaned version of Fashion144K dataset \cite{SimoSerraCVPR2016} contains $90,000$ images. It has multi-label annotations available. There is no similar pair annotation available in the dataset. Along with multiple labels, each image has a fashionability score. The total number of labels are $128$ comprising of parts and color tags. Resolution of images is $384 \times 256$. We use this dataset for discriminative feature learning due to the availability of multiple labels. Fashion550k \cite{InoueICCVW2017} has similar properties as that of its predecessor, Fashion144k \cite{SimoSerraCVPR2016}. Size is four times that of Fashion144k. The total number of labels are $66$.

DeepFashion dataset \cite{LiuCVPR2016} contains $800,000$ images. Similar pair information along with bounding box is available for both consumer-to-shop and in-shop retrieval tasks. For consumer-to-shop retrieval task a total of $33,841$ clothing items are available in $239,557$ consumer/shop image pairs. For in-shop retrieval task $7,982$ clothing item are available in $52,712$ in-shop images.

\section{Experiments and Results}

All the experiments have been conducted on a workstation with 1.2 GHz CPU, 32 GB RAM, NVIDIA P5000 GPU and running Ubuntu 14.04. We use pytorch for the network implementation. We train our model on Fashion144k dataset \cite{SimoSerraCVPR2016}, with 59 item labels, we exclude the color labels for training.

Though not the focus of the work, to confirm the hypothesis that our network is able to detect different clothing regions, we first evaluate the proposed architecture on item recognition task on Fashion144k \cite{SimoSerraCVPR2016} and Fashion550k dataset \cite{InoueICCVW2017}, In these experiments, we use Adam optimizer with a batch size of $64$, momentum of $0.9$ and learning rate of $0.00001$ and trained the models for $40$ epochs. $32$ codewords are used in texture encoding layer. For each image, we assign top $6$ highest-ranked labels to the image and compare with the ground-truth labels. As evaluation metrics for the item recognition task, the class-agnostic average precision (AP\textsubscript{all}), and the mean of each class-average precision (mAP) are used. Table \ref{table:multi-label} shows the comparison.

We also tested our model for item retrieval tasks, for in-Shop as well as consumer-to-shop retrieval tasks on DeepFashion dataset \cite{LiuCVPR2016}. For this task, we extract texture features from the model trained on Fashion144k dataset \cite{SimoSerraCVPR2016} and use Euclidean distance to find similarity between two image features. Fig. \ref{fig:retrieval_results} show the top-k retrieval accuracy of all the compared methods with $k$ ranging from 1 to 50. We also list the top-20 retrieval accuracy after the name of each method.  For in-shop retrieval, our model achieves best performance(0.784) which is better than state-of-the-art FashionNet \cite{LiuCVPR2016} accuracy (0.764). For consumer-to-shop, the corresponding numbers are 0.253 for our model and 0.188 for FashionNet \cite{LiuCVPR2016}.

Finally, we test our model for similar item retrieval task. Fig. \ref{fig:results} shows the retrieval results for various query images. It is easy to see that recommended results have lots of variation in terms of similarity of various parts.

\section{Conclusion}

We have presented a novel attention based deep neural network which learns to attend to different semantic parts using the weakly labeled data. Unlike state of the art techniques using global similarity features, we suggest to extract per part fine-grained features using texture cues: we use DeepTEN in our experiments. We have conducted experiments on three widely accepted tasks in clothing recognition and retrieval. We show that the proposed model outperforms previous state-of-the-art models on deep fashion dataset for both consumer-to-shop and in-shop image retrieval tasks. Our experiment on recommendation task shows that the proposed model is able to successfully recommend a wide variety of objects matching with different parts of the query image.

\bibliographystyle{IEEEbib}
\bibliography{ms}
\end{document}